# Robots Autonomously Detecting People: A Multimodal Deep Contrastive Learning Method Robust to Intraclass Variations

Angus Fung, Beno Benhabib, and Goldie Nejat, *Member, IEEE*

*Abstract*—Robotic detection of people in crowded and/or cluttered human-centered environments including hospitals, stores and airports is challenging as people can become occluded by other people or objects, and deform due to clothing or pose variations. There can also be loss of discriminative visual features due to poor lighting. In this paper, we present a novel multimodal person detection architecture to address the mobile robot problem of person detection under intraclass variations. We present a two-stage training approach using: 1) a unique pretraining method we define as Temporal Invariant Multimodal Contrastive Learning (TimCLR), and 2) a Multimodal YOLOv4 (MYOLOv4) detector for finetuning. TimCLR learns person representations that are invariant under intraclass variations through unsupervised learning. Our approach is unique in that it generates image pairs from natural variations within multimodal image sequences and contrasts crossmodal features to transfer invariances between different modalities. These pretrained features are used by the MYOLOv4 detector for finetuning and person detection from RGB-D images. Extensive experiments validate the performance of our DL architecture in both human-centered crowded and cluttered environments. Results show that our method outperforms existing unimodal and multimodal person detection approaches in detection accuracy when considering body occlusions and pose deformations in different lighting.

## I. INTRODUCTION

Robots need to be able to autonomously detect multiple people in various human-centered environments to engage in effective human-robot interactions. Namely, person detection applications range from long-term care, retirement and private home settings, where interactive robots search for users to provide assistance with activities of daily living [1], product searches in retail stores [2], [3], direction guidance in airports [4] and hospitals [5], [6], and search for victims in urban search and rescue environments (USAR) [7].

In general, human-centered environments can be crowded and cluttered with multiple dynamic people and objects, resulting in person and body part occlusions [7]. Furthermore, as people move or interact in the environment, they can undergo deformation due to both variations in clothing and body articulation [8]. These environments can also have variable illumination due to both natural and artificial lighting sources [9], which can result in appearance changes despite intrinsic properties of the person (*e.g.*, shape) not changing [9]. These variations can be defined as intraclass variations.

Classical learning approaches have been used to detect people by a mobile robot [10]–[13] by extracting a set of expert handcrafted features, such as HOG features [10]–[13], from upright people [14]. However, people exhibit a variety of poses including sitting, lying down, etc. Deep learning (DL) approaches address the limitations of classical learning methods by autonomously learning feature extraction using convolutional neural networks (CNN), without having human experts extract handcrafted features. Thus, they can generalize to people in different poses and postures [15].

To-date, robots use off-the-shelf DL *object detectors* to detect people in indoor environments. The methods include: 1) You Only Look Once (YOLO) [16] used in [17], 2) Single Shot MultiBox Detector (SSD) [18] used in [19], 3) RetinaNet [20] used in [7], and 4) Faster R-CNN (FRCNN) [21] used in [5], [6]. These methods detect people using *unimodal* CNNs from RGB images taken from a single camera on a robot. However, they have difficulty in cluttered environments with varying illumination, as visual features necessary for discriminating people from their backgrounds become less prominent. Furthermore, these DL methods are limited to unimodal CNNs as off-the-shelf *multimodal* pretrained weights do not exist [22], [23].

To address intraclass variations in DL person detection methods, data augmentation has been used to increase the training data [7]. However, this cannot capture the majority of variations due to low probability of occurrence [24]. Unsupervised contrastive learning (CL) can be used to address intraclass variations by pretraining a multimodal model to learn invariant features from scratch and from unlabeled data [25]. CL approaches have been used to learn representation invariances by contrasting between images of different viewpoints of the same static objects in constant lighting conditions [26]. Thus, they have the potential to learn representations which are invariant to intraclass variations. Unlabelled data is relatively inexpensive as a robot can be deployed to autonomously collect multimodal data directly from human-centered environments without the need for manual labelling. Recently, CL methods have been used in a handful of robotic applications [26]–[28]. However, to-date, CL has not been applied to robotic person detection.

In this paper, we present a novel multimodal DL person detection architecture for mobile robots which uses a two-stage training approach consisting of: 1) Temporal Invariant Multimodal Contrastive Learning (*TimCLR*) for pretraining, and 2) Multimodal YOLOv4 (*MYOLOv4*) for finetuning. For prediction, the trained *MYOLOv4* detector is used for autonomous person detection from RGB-D data. We have

Manuscript received: November 24, 2022; accepted April 16, 2023. This paper was recommended for publication by Editor G. Venture upon evaluation of the Associate Editor and Reviewers' comments. This research is supported by the Natural Sciences and Engineering Research Council of Canada (NSERC), HeRo NSERC CREATE program, AGE-WELL Inc., and the Canada Research Chairs (CRC) program. The authors are with the Autonomous Systems and Biomechatronics Lab in the Department of Mechanical and Industrial Engineering at the University of Toronto, 5 King's College Road, Toronto, ON, M5S 3G8 Canada. (Email: angus.fung@mail.utoronto.ca, {benhabib, nejat}@mie.utoronto.ca).

Digital Object Identifier (DOI): see top of this page.

developed a new pretraining method, *TimCLR,* to pretrain a multimodal CNN model from unlabelled RGB-D data in human-centered environments. *TimCLR* incorporates intraclass variations by generating multimodal image pairs from sampling video frames within a short temporal interval, and contrasting person representations within and between modalities, in addition to augmented data. This captures the natural variations in appearance (lighting, occlusions and pose deformations) as people move in their environments. Our overall approach is unique in that it uses CL to consider natural variations in the environment obtained from multimodal features, as well as uniquely incorporates CL within a fusion backbone network to contrast multimodal features. Thus, our approach does not require pretrained RGB weights or expert handcrafted mappings. We present extensive experiments to verify that our DL architecture outperforms existing DL detection methods in human-centered crowded (with dynamic people) and cluttered (with objects) environments under varying lighting conditions.

## II. Related Works

In this section, we discuss the existing DL methods developed for robots to detect multiple dynamic people in human-centered environments and we further introduce CL methods and their current robotic applications.

### A. Person Detection by Robots

Existing person detection methods for robotic applications in indoor environments consist of: 1) single-stage detectors including YOLO [7], [17], [29], [30], SSD [7], [19], and RetinaNet [7]; and 2) two-stage detectors including FRCNN [5]–[7]. In [17], [29], [30], YOLO was used with RGB images for a robot to detect and follow a person to provide assistance. In [19], an SSD detector was used for the same task. All detectors were initialized using off-the-shelf RGB weights which were pretrained on ImageNet [31]. In [5], [6], FRCNN was used to find people with mobility aids in populated environments by a robot using RGB and depth networks which were pretrained on ImageNet. As off-the-shelf RGB weights require 3 input channels, depth images were preprocessed using ColorJet. In our prior work [7], detectors were compared for person/body part detection in cluttered USAR settings from RGB-D images. All the detectors used off-the-shelf RGB weights and were trained by compressing the RGB-D image from 4 to 3 channels with a handcrafted method.

### B. Contrastive Learning Methods

Contrastive learning methods learn representations through similarities/dissimilarities between pairs of images without the need for expert labels [32]. CL methods can be image-based [25], [28], [33] or video-based [26], [27] input type. Image-based methods generate views by applying random data augmentation to the same image [32]. Video-based methods generate diverse views using natural transformations from frames in a video sequence [26] or from various video sequences [27]. The image-based MoCo v3 [25] has been applied to object detection, which generates image pairs by applying data augmentations twice on the same RGB image [25]. MoCo v3 has higher detection accuracy than other image-based methods, including supervised methods [25], [33]. Thus, it has the potential to be applied to robotic person detection. However, it does not consider temporal variations.

CL methods have been used in a handful of robotic applications [26]–[28]. In [26], a video-based CL method learned robot manipulation behaviors by imitating human interactions from videos in an indoor room with constant lighting. A network was pretrained to learn viewpoint-invariant features from RGB images by contrasting between viewpoints of the same scene from different video sequences. In [27], a video-based CL method was used for object discovery by a robot to learn representations of unseen objects in a constant lighting room. This included recognizing and matching an unknown object seen previously to learn viewpoint invariant features from RGB videos. In [28], an image-based CL method was used for robot navigation in a smoked filled environment. CL was applied to learn smoke-invariant representations from LIDAR and radar.

Multimodal CL image-based methods have been used in a handful of papers [35]-[38]. In [35], an image-based RGB-D CL method was proposed to improve representation learning. Negative RGB-D image pairs were generated by replacing the RGB or depth image of a positive pair with that of another pair from a different scene. In [36], an RGB-D CL method was proposed for scene understanding using 3D point clouds and RGB images. Negative point-pixel sample pairs were generated by swapping pixels between positive pairs from different scenes. In [37], an image-based method was proposed for applications where the number of sensors used could change between training and deployment. An RGB and depth network with shared weights was trained using a multiscale CL method [39]. During deployment, either RGB or depth images were used for object recognition. In [38], an image-based CL method was proposed for multimodal image registration, which enforced rotational equivariance by incorporating a constraint into the objective function.

### C. Summary of Limitations

Existing robotic person detection methods have used unimodal CNNs with off-the-shelf RGB weights pretrained on ImageNet, except in [7]. RGB-only approaches have difficulties detecting people under poor lighting due to underexposure [40]. While [7] incorporates RGB-D data, it uses pretrained RGB weights with handcrafted heuristics. To avoid this, off-the-shelf RGB-D weights are required, which do not exist. Although an alternative is to train from scratch using supervised DL methods, RGB-D robotic datasets [6], [7], [41] with detection labels are small, especially compared to ImageNet used during pretraining, or MS COCO used to train RGB detectors from scratch. This can result in overfitting [42]. Instead of using off-the-shelf weights, handcrafted heuristics, or large amounts of annotated data, CL can be used to pretrain RGB-D models from unlabeled data.

Image-based CL methods do not consider intraclass variations. Video-based CL methods, used in a handful of robotic applications, consider temporal variations within RGB images, but have only been applied to learn viewpoint-invariant features of static objects in constant lighting environments [26], [27]. Moreover, video-based methods have not incorporated data augmentation, thereby resulting in feature suppression which degrades representation quality

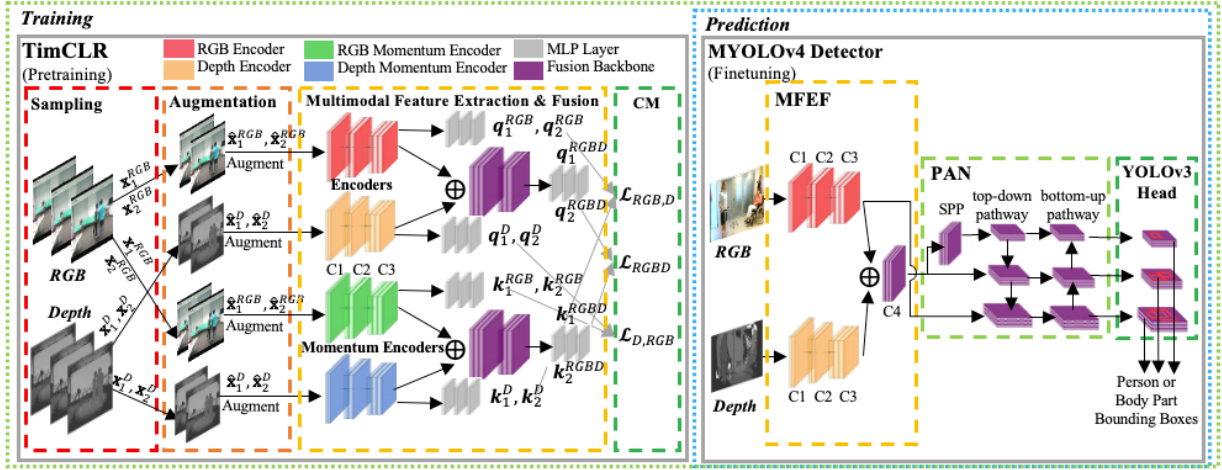

Fig. 1: Proposed multimodal DL detection architecture with first stage *TimCLR* and second stage *MYOLOv4* detector.

[43]. Alternatively, multimodal CL methods [35]–[38] are image-based methods, consisting of separate backbones for each individual modality. However, these methods cannot be directly applied to pretrain our detector which incorporates a fusion backbone network to effectively utilize multiple modalities for person representations. To address the above limitations, we present *TimCLR*, a CL method to generate image pairs by uniquely combining both data augmentations and temporal variations from multimodal images within a short temporal interval. In addition to contrasting features between modalities to transfer invariances learned from one modality to another, we use a fusion backbone to contrast multimodal features. We incorporate this as our novel first stage *TimCLR* pretraining method, whose weights are then transferred to the second stage *MYOLOv4* detector for finetuning and detecting people in RGB-D images.

### III. PERSON DETECTION METHODOLOGY

We propose a person detection architecture to detect multiple dynamic people or body parts from RGB-D images in human-centered environments. The proposed architecture, Fig. 1, consists of two training stages: 1) a *TimCLR* stage, using CL for unsupervised pretraining to learn RGB-D person and body part representations which are robust to intraclass variations, and 2) a *MYOLOv4* stage for supervised finetuning. For prediction, RGB-D images are passed into *MYOLOv4* to detect multiple people/body parts. We selected the state-of-the-art (SOTA) YOLOv4 as it incorporates Path Aggregation Network (PAN) to detect people/body parts at multiple scales [44], which can improve detection under occlusions.

The *TimCLR* stage uses an unlabeled sequence of RGB-D images as inputs. RGB-D image pairs are generated by the *Sampling* module by sampling frames within a short temporal interval. These pairs capture natural variations in an environment by considering similar scenes under different conditions. The *Augmentation* module applies data augmentation to each RGB-D image in the image pair. The *Multimodal Feature Extraction & Fusion* (*MFEF*) module passes RGB-D images into the encoders to extract RGB, depth, and RGB-D person representations. The *Crossmodal (CM)* module maximizes the contrastive loss of those representations generated by the unimodal and fusion backbones. *TimCLR* weights are transferred to the next stage.

The *MYOLOv4* stage uses labelled RGB-D images and the pretrained weights from *TimCLR* for finetuning. *MYOLOv4* adopts the YOLOv4 structure of a PAN and *YOLOv3 head* [44], Fig. 1. The RGB backbone is replaced with a subset of the *TimCLR* backbone layers. The weights from *TimCLR* are used to initialize *MYOLOv4* for training. *MYOLOv4* outputs a bounding box for each detected person within the image.

#### A. Temporal Invariant Multimodal Contrastive Learning

*TimCLR* provides different views by combining both data augmentation and natural variations in the environment obtained from a sequence of RGB and depth images. The main modules of *TimCLR* are discussed below.

*1) Sampling*

The *Sampling* module samples pairs of RGB and depth images from a multimodal dataset $\mathcal{D}$, consisting of sequences of unlabeled images containing people performing activities in a human-centered environment. These sequences capture people (and body parts) undergoing natural variations in occlusion, pose deformation, and lighting. Image pairs are sampled within a short temporal interval $\Delta_t$. Let $(\mathbf{x}_1^{RGB}, \mathbf{x}_1^D)$ and $(\mathbf{x}_2^{RGB}, \mathbf{x}_2^D)$ be RGB-D images sampled at times $t_1$ and $t_2$, an image pair is represented as:

$$\left((\mathbf{x}_1^{RGB}, \mathbf{x}_1^D), (\mathbf{x}_2^{RGB}, \mathbf{x}_2^D)\right) \sim \mathcal{D} \times \mathcal{D}. \quad (1)$$

*2) Augmentation*

The *Augmentation* module applies a set of transformations to each sample pair of images from Eq. (1), which relate two multimodal views representing the same people under different conditions. Namely, the following MoCo v3 transformations are randomly applied [25]: cropping, color jittering, grayscaling, gaussian blurring, and horizontal flipping. Let $\mathbf{T}_1, \mathbf{T}_2 \sim \mathcal{T}$ be the composite of those transformations, and $\hat{\mathbf{x}}_1$ and $\hat{\mathbf{x}}_2$ be the transformed images:

$$(\hat{\mathbf{x}}_1^{RGB}, \hat{\mathbf{x}}_1^D) = \mathbf{T}_1(\mathbf{x}_1^{RGB}, \mathbf{x}_1^D), \quad (2)$$
$$(\hat{\mathbf{x}}_2^{RGB}, \hat{\mathbf{x}}_2^D) = \mathbf{T}_2(\mathbf{x}_2^{RGB}, \mathbf{x}_2^D). \quad (3)$$

The output RGB-D image pairs $\left((\hat{\mathbf{x}}_1^{RGB}, \hat{\mathbf{x}}_1^D), (\hat{\mathbf{x}}_2^{RGB}, \hat{\mathbf{x}}_2^D)\right)$ represent two augmented views of people under different natural variations.

*3) Multimodal Feature Extraction & Fusion*

The *MFEF* module extracts and fuses features from the pairs of RGB-D transformed views to produce RGB, depth, and RGB-D person representations. It consists of an encoder $\mathbf{f}_q$, and momentum encoder $\mathbf{f}_k$ [25]. Each RGB-D image pair is passed into both networks to extract three representations:

$$q_i^{RGBD}, q_i^{RGB}, q_i^D = \mathbf{f}_q(\hat{\mathbf{x}}_i^{RGB}, \hat{\mathbf{x}}_i^D; \boldsymbol{\theta}_q), \qquad i \in \{1,2\} \quad (4)$$

$$k_i^{RGBD}, k_i^{RGB}, k_i^D = \mathbf{f}_k(\hat{\mathbf{x}}_i^{RGB}, \hat{\mathbf{x}}_i^D; \boldsymbol{\theta}_k), \qquad i \in \{1,2\} \quad (5)$$

where $\boldsymbol{\theta}_q$ and $\boldsymbol{\theta}_k$ are the weights of the network, and $q_i^{RGBD}, q_i^{RGB}, q_i^D$ and $k_i^{RGBD}, k_i^{RGB}, k_i^D$ are the feature representations of view $i$ of each modality for the encoder, and momentum encoder, respectively [25]. The encoder weights $\boldsymbol{\theta}_q$ are updated by back-propagation [25]. The momentum weights $\boldsymbol{\theta}_k$ are updated by a weighted average of $\boldsymbol{\theta}_q$ and $\boldsymbol{\theta}_k$ [25], where $m$ is the momentum coefficient:

$$\boldsymbol{\theta}_k \leftarrow m\,\boldsymbol{\theta}_k + (1-m)\,\boldsymbol{\theta}_q. \quad (6)$$

Each encoder consists of separate RGB and depth backbones, fusion backbones, and multilayer perceptrons (MLPs), with weights $\boldsymbol{\theta}_l^{RGB}, \boldsymbol{\theta}_l^D, \boldsymbol{\theta}_l^{RGBD}, \boldsymbol{\theta}_l^{MLP} \in \boldsymbol{\theta}_l, l \in \{q,k\}$, respectively. Each backbone uses a modified ResNet-18 model. The standard RGB ResNet-18 consists of 5 convolution blocks (C1-C5), fully connected (FC), and a SoftMax layer [45]. Our RGB and depth backbones extract RGB and depth features each consisting of C1-C3 blocks, Fig. 1. A fusion backbone is used to concatenate the feature maps of each of the C3 blocks, followed by a 1x1 convolution layer, and C4-C5, Fig. 1. MLPs, consisting of two FC layers, are added to the output of the fusion backbone to extract RGB-D representations. Additional MLPs are added to the output of the RGB and depth backbone layers to extract unimodal representations, Fig. 1. The output representations from the encoders $(q_i^{RGBD}, q_i^{RGB}, q_i^D)$, and $(k_i^{RGBD}, k_i^{RGB}, k_i^D)$, $i \in \{1,2\}$ are passed to the *Crossmodal* module.

*4) Crossmodal*

The *Crossmodal* module computes representation similarity scores using the contrastive loss, $\mathcal{L}_{CL}$, based on InfoNCE [25]:

$$\mathcal{L}_{CL}(\boldsymbol{q},\boldsymbol{k}) = \mathbb{E}_Q\left[\log\frac{\exp(\boldsymbol{q}\cdot\boldsymbol{k}^+/\tau)}{\exp(\boldsymbol{q}\cdot\boldsymbol{k}^+/\tau) + \sum_{\boldsymbol{k}^-}\exp(\boldsymbol{q}\cdot\boldsymbol{k}^-/\tau)}\right], \quad (7)$$

where $Q = \{\boldsymbol{q}_1, \dots, \boldsymbol{q}_N\}$ is the set of representations from the mini-batch; $\{\boldsymbol{k}_1^+, \dots, \boldsymbol{k}_N^+\}$ and $\{\boldsymbol{k}_1^-, \dots, \boldsymbol{k}_N^-\}$ are the set of representations corresponding to positive and negative image pairs; and $\tau$ is the temperature [25]. The RGB-D contrastive loss to measure similarity between representations is:

$$\mathcal{L}_{RGBD} = \mathcal{L}_{CL}(\boldsymbol{q}_1^{RGBD}, \boldsymbol{k}_2^{RGBD}) + \mathcal{L}_{CL}(\boldsymbol{q}_2^{RGBD}, \boldsymbol{k}_1^{RGBD}). \quad (8)$$

The crossmodal contrastive losses to measure similarity between unimodal RGB and depth representations are:

$$\mathcal{L}_{RGB,D} = \mathcal{L}_{CL}(\boldsymbol{q}_1^{RGB}, \boldsymbol{k}_2^D) + \mathcal{L}_{CL}(\boldsymbol{q}_2^{RGB}, \boldsymbol{k}_1^D), \quad (9)$$

$$\mathcal{L}_{D,RGB} = \mathcal{L}_{CL}(\boldsymbol{q}_1^D, \boldsymbol{k}_2^{RGB}) + \mathcal{L}_{CL}(\boldsymbol{q}_2^D, \boldsymbol{k}_1^{RGB}). \quad (10)$$

The full contrastive loss which is the combination of all the aforementioned losses is defined as:

$$\mathcal{L}_{MCL} = \lambda_{RGBD}\mathcal{L}_{RGBD} + \lambda_{RGB,D}\mathcal{L}_{RGB,D} + \lambda_{D,RGB}\mathcal{L}_{D,RGB}, \quad (11)$$

where $\lambda_{RGBD}, \lambda_{RGB,D}$, and $\lambda_{D,RGB}$ are the weighting factors. The encoder weights $\boldsymbol{\theta}_q$ are passed to the *MYOLOv4* stage.

### B. Multimodal YOLOv4

*MYOLOv4*, Fig. 1, adopts its structure from YOLOv4 [44]. It consists of *MFEF*, *PAN*, and *YOLOv4 head* modules. *MFEF* extracts feature maps from RGB and depth images using separate RGB and depth backbones each consisting of C1-C3 blocks which are initialized by $\boldsymbol{\theta}_q^{RGB}$ and $\boldsymbol{\theta}_q^D$, the unimodal backbones weights in *TimCLR*. These backbones are fused by concatenation with a 1x1 convolution layer, followed by a C4 block, which is initialized by $\boldsymbol{\theta}_q^{RGBD}$, the fusion backbone weights in *TimCLR*. The *PAN* module uses C3 and C4, and consists of a Spatial Pyramid Pooling (SPP) block [44], and top-down/bottom-up pathways to aggregate features at different scales. SPP, C4, and C3 are concatenated after each upsampling layer, respectively. The bottom-up pathway consists of convolutions to produce feature maps of the same scales, with lateral connections to concatenate the corresponding top-down feature map. The *YOLOv3 head* module uses these feature maps to predict people/body parts at each scale. *MYOLOv4* is first trained through supervised learning and used during prediction to output a bounding box for each detected person or body parts within an image.

## IV. DATASETS

We used a number of datasets to train our architecture for mobile robot person detection under intraclass variations:
1) **NTU RGB+D 120 dataset** [46] which consists of multiple people performing 120 different actions (*e.g.,* standing, eating, jumping) in indoor environments with 114,480 RGB-D video samples collected by a Kinect sensor. NTU naturally captures the intraclass variations that would be expected in a real-world human-centered environment.
2) **MS COCO dataset** [34] which consists of 123,000 RGB images of general objects with 250,000 person instances in indoor/outdoor environments. The entire dataset was used for training to learn semantically rich RGB person features.
3) **MA dataset [5]** which consists of 17,000 annotated RGB-D images of multiple dynamic people undergoing frequent occlusions in a crowded real-world hospital environment, collected by a Kinect sensor on a robot. The dataset contains 5 classes of people with different aids. We have relabeled and combined these multiple classes into a generalized single person class. The dataset was split 65/35 for training/testing. The test sets (TS), from [5], are: 1) TS1 (few people occlusions), and 2) TS2 (frequent people occlusions).
4) **USAR dataset** [7] which consists of 570 RGB-D images of human/mannequin body parts in a real-world cluttered environment, collected by a Kinect sensor on a Turtlebot 2 robot. The dataset contains 6 classes: arm, foot, hand, head, leg, and torso; split using an 80/20 rule, with 3 test datasets: 1) TS1 (contains fully visible people), 2) TS2 (contains partial person or body part occlusions and deformations), and 3) TS3 (contains people under low lighting conditions).
5) **IOD dataset** [47] which consists of 8,300 annotated RGB-D images of multiple dynamic people under changing lighting conditions in both indoor and outdoor university campus environments, collected by a Kinect sensor on a robot. The dataset contains a single person class.

## V. TRAINING

The proposed person detection architecture was trained in two stages: 1) *TimCLR*, and 2) *MYOLOv4*. For *TimCLR* pretraining, a subset of the unlabeled NTU dataset of 1 million RGB-D images was used, generated by randomly selecting from the 114,480 videos in the dataset. Image pairs were sampled at $\Delta_t = 50$ frames. *TimCLR* was trained for 100 epochs. For finetuning *MYOLOv4*, the RGB branch of *MYOLOv4* was first finetuned using the *TimCLR* weights on the MS COCO dataset. For evaluation on the MA, USAR, and IOD test sets, *MYOLOv4* was additionally finetuned only on the corresponding training set using stochastic gradient descent with a learning rate of 0.01 for 26 epochs.

VI. EXPERIMENTS

Our proposed two-stage person detection architecture is investigated in three sets of experiments: 1) a comparison study with existing robot person detection methods to evaluate detection accuracy, 2) an ablation study to validate the design choices of our architecture, and 3) a comparison study with contrastive learning methods. Three environments with varying levels of person occlusion, illumination, and deformation were considered, from which RGB-D images have been taken by a mobile robot. All experiments were trained on a workstation with two RTX 3070 GPUs, an AMD Ryzen Threadripper 3960X, and 128GB of memory.

*A. Performance Metrics*

The mean average precision (mAP) was chosen as the performance metric for detection accuracy. $AP_{50}$ is the mAP where predictions with an Intersection over Union (IoU) > 0.5 are considered true positives. $AP_{0.5:0.95}$, or AP, is the averaged mAP over IoU = {0.5, 0.55, ..., 0.95}, which more accurately measures localization [34], and is the primary accuracy metric. We also measured the memory usage and inference speed in frames per second (FPS) on an Nvidia Jetson AGX Xavier platform using the TensorRT framework.

*B. Comparison Methods*

Our person detection method was compared against existing DL RGB person detection methods, including RGB: 1) SSD-300 [18], 2) RetinaNet-FPN [20], 3) FRCNN [21], 4) YOLOv4 [44], and 5) EfficientDet-D0 [48]; 6) depth ColorJet (CJ) FRCNN [6] which uses CJ to preprocess depth images into 3-channels; and 7) RGB-D Compression (C) FRCNN [7] which uses compression to preprocess RGB-D images. Please note that distinct from [6], we are optimizing the mAP for detecting individual people in a scene rather than five specific different mobility aids. We also designed the following RGB-D strong baselines: 8) CJ-MYOLOv4, 9) CJ Multimodal FRCNN (CJ-MFRCNN), and 10) CJ Multimodal EfficientDet (CJ-MEfficientDet). We also compare against RGB-D TimCLR + MFRCNN, which uses our *TimCLR* method with the two-stage detector FRCNN which has been found to be accurate in detecting people in both crowded and cluttered environments [49], [50]. By comparing with RGB-D TimCLR + MFRCNN, we investigate the ability of *TimCLR* to learn invariant person features regardless of the specific finetuning method used. Finally, we compare against the top four SOTA methods which obtain the highest AP from the literature reported on the IOD dataset: 1) RGB-D Mixture of Deep network Experts (MoDE) detector [47], 2) depth-based Fast Region Proposal Generation with FRCNN (Depth FRPG-FRCNN) [5], 3) RGB-D FRCNN Y-fusion [51], and 4) RGB-D FRCNN U-fusion detector [51]. The MoDE detector uses a mixture of expert CNNs for RGB, depth, and optical flow, with a gating network to weigh the classifier outputs of each expert. The Depth FRPG-FRCNN detector uses a depth-based method to generate initial proposals from point clouds using Euclidean clustering, followed by FRCNN for detecting people. RGB-D FRCNN Y-fusion and U-fusion detectors incorporate FRCNN to detect people by fusing RGB and depth backbone feature outputs and classifier outputs. Off-the-shelf ImageNet pretrained weights were used.

We pretrained the networks on the NTU dataset, the same dataset used to train *TimCLR*, to ensure comparison fairness. To serve as strong baselines, we also pretrained with the unsupervised MoCo v3, as the NTU dataset (unlike ImageNet), does not provide classification labels. The exception being TimCLR + MFRCNN which is pretrained using the procedure in Section V. All networks were pretrained using ResNet-18 for 100 epochs, except for the EfficientDet methods which used EfficientNet-B0 backbone [48]. We used the smaller ResNet with 18 layers to meet the inference requirements of robotic applications. The networks were finetuned using the procedure from Section V.

*C. People Detection Comparison Results*

The detection accuracy results for our proposed method and all comparison methods are presented in Table I. Our proposed *TimCLR + MYOLOv4* detector outperformed the other methods with respect to AP and $AP_{50}$ on all test sets (TS), including on 1) partial occlusions (MA/USAR TS2), 2) deformations (USAR TS2), and 3) varying illuminations (USAR TS3, IOD TS). The results show *TimCLR*'s ability to learn invariant person features. Namely, our method outperformed all the RGB-only networks. While the RGB approaches generally outperformed the depth-only methods, they performed worse under varying lighting (USAR TS3).

Worth noting is that our proposed method outperformed the strong multimodal baselines that we designed which were all pretrained on MoCo v3. This further highlights the direct performance benefits due to *TimCLR*. For example, *TimCLR + MYOLOv4* outperformed RGB-D CJ-MYOLOv4 with 11% and 22% improvements in AP under partial occlusions on the MA and USAR datasets, respectively. Under varying lighting, it achieved a 5% and 10% improvement in AP on the IOD and USAR datasets. These results are statistically significant, $p < 0.001$. In general, *TimCLR* was more effective than CJ at capturing depth features for differentiating people/body parts from background clutter (USAR TS2) and under poor lighting (USAR TS3, IOD TS). Compared to RGB-D C-FRCNN, our method had improvements of 16-40% on AP across all test sets. We postulate that RGB-D C-FRCNN had difficulties in estimating the extent of a person/body part under occlusions, as visual features may be lost during compression. Thus, our method accurately localizes multiple people/body parts with a greater overlap between the predictions and ground truth.

Furthermore, our method outperformed RGB-D TimCLR + MFRCNN across all test sets, despite the latter incorporating a two-stage detector. This is a result of *PAN* in YOLOv4 which allows for detection of body parts at multiple scales. Thus, our method is more effective at detecting body parts when discriminative features occupy smaller regions due to occlusions or poor lighting, as seen by larger improvements in AP on TS2-3 compared to TS1. Overall, our *TimCLR*-based methods were more accurate and robust in cluttered/crowded environments. Thus, *TimCLR* can be

TABLE I  Comparison of Detection Accuracy of our Proposed Detection Method versus Existing Detection Methods

| Dataset | Mobility Aids (MA) | | | | USAR | | | | | | IOD | | Mem (GB) | FPS |
|---|---|---|---|---|---|---|---|---|---|---|---|---|---|---|
| | Test Set 1 | | Test Set 2 Occlusion | | Test Set 1 | | Test Set 2 Occlusion + Deformation | | Test Set 3 Illumination | | Test Set | | | |
| Method | AP | $AP_{50}$ | AP | $AP_{50}$ | AP | $AP_{50}$ | AP | $AP_{50}$ | AP | $AP_{50}$ | AP | $AP_{50}$ | | |
| RGB-D *TimCLR + MYOLOv4* | 60.10 | 94.30 | 49.20 | 75.30 | 22.30 | 45.80 | 21.10 | 49.60 | 20.20 | 44.40 | 63.40 | 96.70 | 1.4 | 41 |
| RGB-D *TimCLR* + MFRCNN | 58.81 | 94.05 | 47.23 | 75.41 | 22.02 | 45.80 | 17.95 | 40.29 | 18.60 | 37.07 | 63.12 | 96.44 | 2.1 | 4 |
| RGB YOLOv4 [44] | 55.90 | 87.40 | 44.40 | 70.11 | 15.50 | 35.80 | 14.90 | 34.40 | 15.10 | 34.20 | 60.50 | 91.90 | 1.4 | 42 |
| RGB FRCNN [21] | 55.63 | 92.74 | 44.85 | 73.49 | 14.52 | 37.45 | 15.73 | 35.47 | 10.32 | 29.55 | 49.90 | 85.84 | 2.1 | 4 |
| RGB EfficientDet [48] | 55.52 | 87.26 | 44.66 | 75.41 | 15.36 | 37.78 | 13.71 | 37.36 | 16.11 | 35.85 | 59.70 | 90.86 | 2.0 | 25 |
| RGB SSD [18] | 35.70 | 87.91 | 26.98 | 71.57 | 10.70 | 30.04 | 11.21 | 32.83 | 8.62 | 30.81 | 38.60 | 58.47 | 2.0 | 14 |
| RGB RetinaNet [20] | 52.06 | 93.33 | 42.89 | 75.20 | 13.80 | 34.95 | 14.21 | 33.13 | 13.65 | 28.18 | 55.92 | 85.41 | 1.5 | 19 |
| Depth CJ-FRCNN [6] | 41.36 | 84.27 | 32.95 | 67.28 | 10.98 | 25.76 | 9.87 | 26.70 | 17.63 | 34.84 | 44.64 | 67.97 | 2.1 | 19 |
| RGB-D C-FRCNN [7] | 47.86 | 90.00 | 40.51 | 73.43 | 15.92 | 38.84 | 15.72 | 41.59 | 18.27 | 35.42 | 51.57 | 88.52 | 2.1 | 4 |
| RGB-D CJ-MFRCNN | 56.20 | 92.84 | 45.75 | 70.38 | 20.50 | 44.28 | 17.18 | 41.60 | 18.28 | 35.94 | 55.36 | 92.77 | 2.1 | 4 |
| RGB-D CJ-MYOLOv4 | 57.40 | 89.20 | 45.20 | 71.40 | 20.90 | 44.70 | 17.30 | 41.20 | 18.30 | 40.60 | 60.80 | 92.90 | 1.4 | 41 |
| RGB-D CJ-MEfficientDet | 56.90 | 87.75 | 45.71 | 74.54 | 20.05 | 44.68 | 17.02 | 40.62 | 17.80 | 40.45 | 60.68 | 92.06 | 2.0 | 24 |
| Depth FRPG-FRCNN [5] | - | - | - | - | - | - | - | - | - | - | - | 70.00 | - | 23 |
| RGB-D MoDE [47] | - | - | - | - | - | - | - | - | - | - | - | 80.40 | - | - |
| RGB-D FRCNN Y-fusion [51] | - | - | - | - | - | - | - | - | - | - | - | 90.10 | - | 5 |
| RGB-D FRCNN U-fusion [51] | - | - | - | - | - | - | - | - | - | - | - | 94.40 | - | 3 |

Note: all missing values above (denoted '-') were not reported.

applied to pretrain any SOTA detector with high accuracy.

Our *TimCLR*-based methods also outperform the four SOTA methods on the IOD dataset, with *TimCLR + MYOLOv4* achieving up to 38% improvements in $AP_{50}$. This illustrates the advantages of *TimCLR* for learning lighting invariant features. Fig. 2 presents example detections of our *TimCLR + MYOLOv4* method (Fig. 2(a)) compared to RGB-D CJ-MYOLOv4 (Fig. 2(b)), the best RGB-D baseline; and RGB-D C-FRCNN, the best robotic person detection baseline (Fig. 2(c)). Fig. 2 shows these methods under partial occlusions (rows 1-2), pose deformation (row 2), and poor lighting (row 3), on the MA (row 1), USAR (row 2), and IOD (row 3) datasets. For example, row 2 shows a person partially occluded by rubble and a mannequin in the fetal position, and row 3 shows three people under low lighting. Our method identified all the people in Fig. 2(a) rows 1 and 3; and it was the only method to detect the partially occluded right foot in the second scene. While RGB-D CJ-MYOLOv4 detected a part of the left leg in Fig. 2(b), it did not capture the articulated portion of the leg as our method did.

A non-parametric Kruskal-Wallis test was performed on all the datasets, showing a statistically significant difference in AP between the multimodal methods. A post-hoc Dunn test with Bonferroni correction showed our *TimCLR + MYOLOv4* had a statistically significant higher AP than the alternatives. The statistical test results are provided on our lab website[1]. In general, the YOLOv4 detectors had the lowest memory usage, with the fastest inference of 41-42 FPS. Thus, our *TimCLR + MYOLOv4* is most suited for mobile robot detection of people in human-centered cluttered and crowded environments.

### D. Ablation Study

We performed an ablation study to evaluate the design of our proposed detection architecture. We investigated *TimCLR* with respect to: 1) image pair generation, 2) fusion design, and 3) crossmodal loss. We evaluated each design choice based on the detection accuracy of *MYOLOv4* using the pretrained weights from *TimCLR*, presented in Table II.

In experiment 1, we investigated generating positive image pairs, using: 1) data augmentation, 2) natural variations, and

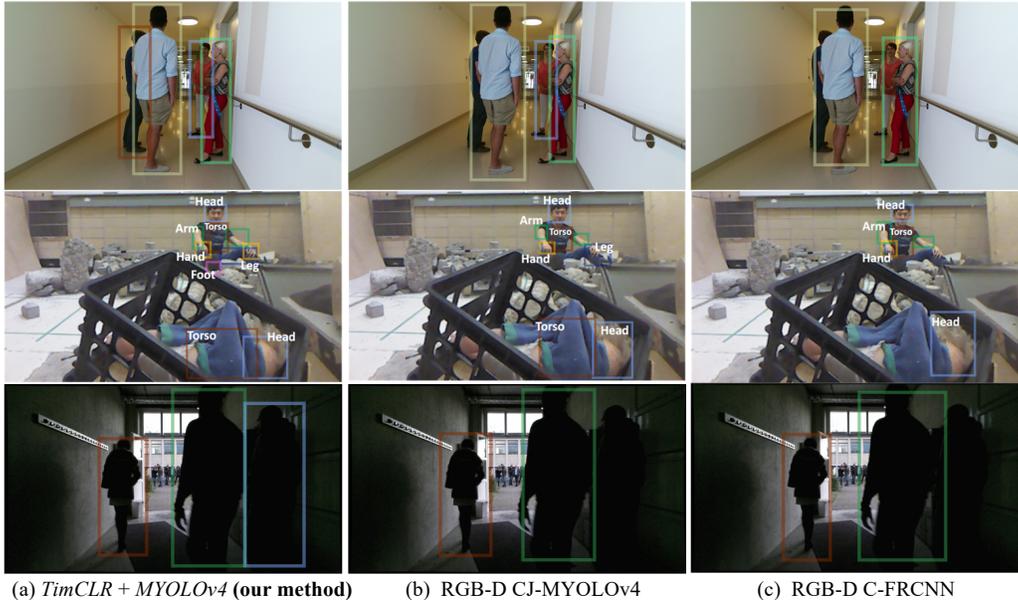

(a) *TimCLR + MYOLOv4* **(our method)**  (b) RGB-D CJ-MYOLOv4  (c) RGB-D C-FRCNN

Fig. 2: Multimodal detection results from: (a) *TimCLR + MYOLOv4* (ours), (b) RGB-D CJ-MYOLOv4, (c) RGB-D C-FRCNN; overlaid on RGB images.

[1]http://asblab.mie.utoronto.ca/sites/default/files/Supplementary%20Material%20TimCLR.pdf

TABLE II ABLATION STUDY

| Dataset | Mobility Aids (MA) | | | | USAR | | | | | | IOD | |
|---|---|---|---|---|---|---|---|---|---|---|---|---|
| | Test Set 1 | | Test Set 2 Occlusion | | Test Set 1 | | Test Set 2 Occlusion+Deformation | | Test Set 3 Illumination | | Test Set | |
| Method | AP | $AP_{50}$ | AP | $AP_{50}$ | AP | $AP_{50}$ | AP | $AP_{50}$ | AP | $AP_{50}$ | AP | $AP_{50}$ |
| *TimCLR*, data augmentation only | 58.10 | 90.40 | 45.30 | 66.40 | 19.80 | 43.90 | 16.90 | 45.10 | 18.00 | 40.00 | 61.70 | 92.50 |
| *TimCLR*, natural variations only | 48.70 | 90.30 | 40.40 | 64.90 | 18.10 | 41.80 | 17.40 | 42.80 | 16.20 | 37.50 | 60.10 | 91.70 |
| *TimCLR*, combined | 60.10 | 94.30 | 49.20 | 75.30 | 22.30 | 45.80 | 21.10 | 49.60 | 20.20 | 44.40 | 63.40 | 96.70 |
| *TimCLR*, C3 Fusion | 60.10 | 94.30 | 49.20 | 75.30 | 22.30 | 45.80 | 21.10 | 49.60 | 20.20 | 44.40 | 63.40 | 96.70 |
| *TimCLR*, C4 Fusion | 57.40 | 90.10 | 46.00 | 65.70 | 20.20 | 43.70 | 19.40 | 48.70 | 18.30 | 41.10 | 60.70 | 93.10 |
| *TimCLR*, C5 Fusion | 54.20 | 87.40 | 44.40 | 65.10 | 18.30 | 40.90 | 18.20 | 43.50 | 17.50 | 40.00 | 58.80 | 91.20 |
| *TimCLR*, no crossmodal loss | 57.80 | 90.30 | 47.70 | 71.10 | 18.40 | 41.70 | 17.50 | 42.70 | 16.20 | 37.80 | 60.00 | 93.90 |
| *TimCLR*, with crossmodal loss | 60.10 | 94.30 | 49.20 | 75.30 | 22.30 | 45.80 | 21.10 | 49.60 | 20.20 | 44.40 | 63.40 | 96.70 |

3) a combination. Pretraining using only data augmentation outperforms using only natural variations on TS1 as a result of feature suppression. However, they perform similarly under intraclass variations (TS2-3). During feature suppression, the network may ignore texture or shape features as other cues (*e.g.*, color distributions) can be used to differentiate between pairs. The combined approach avoids this, achieving higher detection accuracy, while still incorporating natural variations. In experiment 2, we investigated the fusion of RGB and depth features at C3, C4, or C5 encoder blocks. We found that C3 was the optimal layer to fuse. Finally, we investigated the crossmodal contrastive loss which contrasts representations between RGB and depth for evaluating the transfer of knowledge between modalities. We performed runs with and without this loss. The run with the loss performed similarly on TS1, but substantially better under intraclass variations (TS2-3). One possible reason is that contrasting of features between modalities encourages representations between them to be similar, thus transferring learned feature invariances from one modality to the other.

*E. Contrastive Pretraining Comparison Results*

We performed a comparison study to evaluate *TimCLR* with respect to SOTA CL methods such as SimCLR [32], Barlow Twins [33], and MoCo v3 [25] on downstream detection tasks. Specifically, we pretrain the methods on the NTU dataset for 100 epochs, for both the RGB and depth modality. The networks were finetuned using the procedure from Section V. The results are presented in Table III.

Across all datasets, our RGB and depth *TimCLR* + YOLOv4 methods outperformed the other RGB and depth pretraining methods. We observed improvements with our RGB and depth *TimCLR* methods of up to: 1) 21% and 18% in AP under partial occlusions (MA/USAR TS2), 2) 30% and 22% under lighting variations (USAR TS3), and 3) 6% for both RGB and depth on the IOD TS. These improvements under intraclass variations emphasize the advantage of our

*TimCLR* in generating image pairs within a short temporal window to allow for the capturing and learning of features robust to intraclass variations. Non-parametric Kruskal-Wallis and post-hoc Dunn tests performed on all datasets showed that our *TimCLR* + *MYOLOv4* had a statistically significant higher AP than the other CL methods, $p < 0.001$.

VII. DISCUSSIONS

Our proposed *TimCLR* + *MYOLOv4* method outperformed the SOTA on both the person detection (Section VI.C) and contrastive pretraining (Section VI.E) experiments, demonstrating that our approach is effective for addressing the challenge of person detection under intraclass variations in diverse environments. While *TimCLR* can extract intraclass-invariant features by using video sequences for pretraining, *MYOLOv4,* similar to the existing person detection methods in Section IV.C, has been designed to use static images. Thus, it does not incorporate temporal context during real-time person detection and would not be able to handle full person occlusion scenarios. Existing datasets have also mainly focused on pose deformations, with limited instances of clothing deformations. It would be worthwhile to consider creating a dataset that also includes a diverse range of clothing variations for pretraining of our multimodal person detection architecture.

VIII. CONCLUSIONS

In this paper, we present a novel multimodal person detection architecture for mobile robots to address the robotic problem of person detection under intraclass variations. We introduce a new pretraining method *TimCLR* which learns person features which are invariant to natural variations in the environment, such as person and body part occlusions, pose deformations, and varying lighting. Our *TimCLR* generates contrastive image pairs by sampling natural variations from multimodal image sequences within a short temporal interval, in addition to data augmentation. These invariant person features are used by *MYOLOv4* for robust detection of people under intraclass variations. Extensive experiments verified that our *TimCLR + MYOLOv4* outperformed the existing detection methods in finding people in crowded and/or with varying lighting hospitals, university campuses, and cluttered USAR environments. Our ablation study validated the design choices of *TimCLR*, and our comparison study with SOTA CL methods showed that our *TimCLR* is more robust in learning person representations, even in the single-modality case. Future work includes integrating our detection architecture within a mobile robot for real-time person detection in varying human-centered environments.

TABLE III COMPARISON OF CONTRASTIVE PRETRAINING METHODS ON DOWNSTREAM PERSON DETECTION

| Dataset | Mobility Aids (MA) | | | | USAR | | | | | | IOD | |
|---|---|---|---|---|---|---|---|---|---|---|---|---|
| | Test Set 1 | | Test Set 2 Occlusion | | Test Set 1 | | Test Set 2 Occlusion+Deformation | | Test Set 3 Illumination | | Test Set | |
| Method | AP | $AP_{50}$ | AP | $AP_{50}$ | AP | $AP_{50}$ | AP | $AP_{50}$ | AP | $AP_{50}$ | AP | $AP_{50}$ |
| RGB SimCLR + YOLOv4 | 54.10 | 88.10 | 43.40 | 70.20 | 15.30 | 35.00 | 14.60 | 28.10 | 14.50 | 28.30 | 58.30 | 92.40 |
| RGB Barlow Twins + YOLOv4 | 55.40 | 85.50 | 43.60 | 70.30 | 14.90 | 33.70 | 13.90 | 31.60 | 13.70 | 32.10 | 60.50 | 92.10 |
| RGB MoCo v3 + YOLOv4 | 55.90 | 87.40 | 44.40 | 70.11 | 15.50 | 35.80 | 14.90 | 34.40 | 15.10 | 34.20 | 60.50 | 91.90 |
| RGB *TimCLR* + YOLOv4 | 57.90 | 92.50 | 46.10 | 72.20 | 17.60 | 42.20 | 16.80 | 40.10 | 17.80 | 41.10 | 61.60 | 94.20 |
| Depth SimCLR + YOLOv4 | 42.00 | 81.50 | 35.50 | 65.10 | 17.30 | 41.30 | 17.60 | 46.70 | 15.80 | 32.50 | 58.40 | 88.40 |
| Depth Barlow Twins + YOLOv4 | 44.50 | 82.00 | 33.60 | 63.30 | 17.90 | 40.20 | 17.40 | 43.30 | 17.20 | 36.70 | 57.80 | 89.00 |
| Depth MoCo v3 + YOLOv4 | 44.70 | 87.50 | 35.20 | 65.10 | 18.30 | 41.90 | 17.80 | 43.50 | 17.60 | 37.80 | 59.80 | 89.50 |
| Depth *TimCLR* + YOLOv4 | 46.10 | 91.10 | 39.50 | 69.20 | 20.70 | 42.40 | 20.20 | 48.50 | 19.20 | 39.80 | 61.40 | 91.40 |
| RGB-D *TimCLR* + MYOLOv4 | 60.10 | 94.30 | 49.20 | 75.30 | 22.30 | 45.80 | 21.10 | 49.60 | 20.20 | 44.40 | 63.40 | 96.70 |